\documentclass{article}

\usepackage[preprint, nonatbib]{nips_2018}

\usepackage{url}
\usepackage{amsmath}
\usepackage{capt-of}
\usepackage{algorithm}
\usepackage{algorithmic}

\usepackage{graphicx}
\usepackage{lipsum}
\usepackage{amsfonts}
\usepackage{graphicx}
\usepackage{epstopdf}
\usepackage{amsmath}
\usepackage{array}
\usepackage{stfloats}
\usepackage{multicol}
\usepackage{multirow}
\usepackage{mathtools}
\usepackage{amssymb}
\usepackage{makecell}
\usepackage{subfigure}
\usepackage{adjustbox}
\usepackage[numbers]{natbib}

\makeatletter
\newcommand{\thickhline}{%
    \noalign {\ifnum 0=`}\fi \hrule height 1pt
    \futurelet \reserved@a \@xhline
}
\newcolumntype{"}{@{\hskip\tabcolsep\vrule width 1pt\hskip\tabcolsep}}
\makeatother

\title{Sampled Policy Gradient for Learning to Play the Game Agar.io}

\author{
Anton Orell Wiehe \\ Department of Artificial Intelligence \\ Bernoulli Institute \\ University of Groningen \\ \texttt{antonwiehe@gmail.com} \\
\And 
Nil Stolt Ans\'o \\ Department of Artificial Intelligence \\ Bernoulli Institute \\ University of Groningen \\ \texttt{nilstoltanso@gmail.com} \\
\And
Madalina M. Drugan \\ ITLearns.Online \\ Utrecht \\ \texttt{madalina.drugan@gmail.com} \\
\And 
Marco A. Wiering \\ Department of Artificial Intelligence \\ Bernoulli Institute \\ University of Groningen \\ \texttt{m.a.wiering@rug.nl} \\
}

\begin{document}

\maketitle

\begin{abstract}
In this paper, a new offline actor-critic learning algorithm is introduced: Sampled Policy Gradient (SPG). SPG samples in the action space to calculate an approximated policy gradient by using the critic to evaluate the samples. This sampling allows SPG to search the action-Q-value space more globally than deterministic policy gradient (DPG), enabling it to theoretically avoid more local optima. 
SPG is compared to Q-learning and the actor-critic algorithms CACLA and DPG
in a pellet collection task and a self play environment in the game Agar.io. 
The online game Agar.io has become massively popular on the internet due to intuitive game design and the ability to instantly compete against players around the world. From the point of view of artificial intelligence this game is also very intriguing: The game has a continuous input and action space and allows to have diverse agents with complex strategies compete against each other. 
The experimental results show that Q-Learning and CACLA outperform a pre-programmed greedy bot in the pellet collection task, but all algorithms fail to outperform this bot in a fighting scenario.  The SPG algorithm is analyzed to have great extendability through offline exploration and it matches DPG in performance even in its basic form without extensive sampling.

\end{abstract}

\section{Introduction}\label{sec:introduction}

Reinforcement learning (RL) is a machine learning paradigm which uses a reward function that assigns a value to a specific state an agent is in as a supervision signal \cite{RL:AI}. The idea is that the agent finds out what actions to take in an environment to maximize the overall intake of this reward signal. This is often done by having a critic network, which predicts the value of a state or of an action in a state. The value is an estimation of the total expected reward that will be gained in the future. This critic is then used to choose which action is best in a state: the one that has the highest value.

RL agents are usually trained in simulations or games. This is for multiple reasons: the level of noise can be directly controlled, the researcher has access to all relevant information and the simulation can be sped up and parallelized. As algorithms become able to solve more and more complex and noisy simulated environments they can also be employed for real-world tasks.

The environment for this research is inspired by the game of \url{Agar.io}. In this game the player controls circular cells in a 2D plane which follow the player's mouse cursor. The player can split its cells in half or eject mass out of cells. Cells of a player can eat either small food pellets or cells of other smaller (human) online players to grow in size. The environment of Agar.io is very interesting for RL research, as it is mainly formed by other players on the same plane, it is stochastic and constantly changing. Also the output space is continuous, as the cells of the player move towards the exact position of the mouse cursor. On top of that, the complexity of the game can be scaled by introducing or removing additional features. This paper therefore studies how to use RL to train an intelligent agent for this game.

In constructing this agent this research focuses especially on which algorithm to choose to optimize the reward signal in the long run. Along with comparing three conventional algorithms, a new algorithm called sampled policy gradient is introduced. This new algorithm builds a bridge between the already existing algorithms DPG \cite{DPG} and CACLA \cite{CACLA} and therefore enables a better comparison between the two. Furthermore the algorithm shows promising results and can be heavily extended because it can be used with offline exploration.

For most complex problems it is impossible to rely on tabular entries of state values. Therefore RL research often makes use of artificial neural networks, which have been shown to be able to approximate any continuous function to any degree of accuracy \cite{Cybenko1989}. Tesauro was among the first to show that super-human decision making could be learned in the large state space of the game Backgammon through the use of a multi-layer perceptron (MLP) \cite{tesauro1995}. Over the years, this principle has been extended through the use of convolutional neural networks (CNNs). While Tesauro's backgammon agent was still heavily reliant on high-level strategic state features of the game, CNNs allow an agent to abstract its own features from raw sensory input, such as pixels. This has been shown to achieve human level performance on a variety of Atari games \cite{Mnih2013}, and even 3D first-person shooter games like Doom \cite{lample2017}.

Human-engineered vision grids are an intermediate step between raw pixel input and high-level features that require expert knowledge to construct. Such vision grids contain low-level features to encode the state. In research on the game Tron \cite{knegt2017} multiple binary grids were used: one indicating the presence of an opponent, one for the presence of walls, and one for the presence of the trails made by the player itself. These vision grids have been used in more complex environments, such as Starcraft \cite{shantia2011}. In the research on Starcraft the vision grids indicated information such as weapon range, danger and information on the terrain. Similar to the paper on Tron this research on Agar.io will use vision grids to indicate the positions of cells and the wall.

To produce continuous actions an actor network is required, which maps states to actions. This paper will compare two existing methods that make use of such an actor network: continuous actor critic learning automaton (CACLA) \cite{CACLA} and deterministic policy gradient (DPG) \cite{DPG}. Both approaches rely on a critic network, which maps states or state-action pairs to quality values: how much reward is to be expected in the long term. This critic is comparable to the network used in Q-learning and it is trained in a similar manner. The actor training in CACLA analyzes a transition from an action in a state to a produced reward and a new state. If this action positively surprises the critic because it gives an unexpectedly good reward or leads to an unexpectedly good new state, then it will be reinforced in the future by being made more likely to be predicted by the actor network.

The actor training of DPG on the other hand is not dependent on transitions, it only requires a valid state of the environment. By applying backpropagation through the critic network, it is calculated in what direction the action input of the critic needs to change, to maximize the output of the critic. Then DPG tunes the weights of the actor to change its output in this previously determined direction.

\subsection{Contributions of this Paper}
This paper focuses on multiple contributions:
\begin{enumerate}
    \item The construction of a reinforcement learning agent that learns to play the game Agar.io will be investigated. This will be done by assessing what state representation can be used, what reward function is appropriate, and what learning algorithm performs best.
    \item A new continuous-action actor-critic algorithm is introduced: Sampled Policy Gradient (SPG). Also, a first step towards an offline exploration strategy for SPG is investigated.
    \item In many RL problems requiring continuous actions DPG is used. This paper therefore explores whether this is also warranted for the game of Agar.io and compares its performance to CACLA and SPG.
    \item Lastly, it is investigated whether a discretization of the action space enables Q-learning to get to a similar performance as the actor-critic algorithms despite its lack in precision.
\end{enumerate}

\subsection{Outline of this Paper}
Section~\ref{sec:rl} explains the background knowledge on RL. 
Section~\ref{SPG} introduces sampled policy gradient.
Section 4 explains the Agar.io game more in depth together with the 
used state representation. 
Section~\ref{sec:experiments} shows the experimental results on pellet collection and on fighting a preprogrammed bot. 
Section~\ref{sec:conclusions} presents the conclusions and describes some directions for future work.

%%%%%%%
% background ??!!
%\input{tex/rl.tex}
\section{Reinforcement Learning}\label{sec:rl}

This paper follows the general conventions \cite{RL:AI} to model the reinforcement learning (RL) problem as a Markov decision process (MDP). 
In a Markov Decision Process, an agent takes an action in a state to get to a new state, where the transition probabilities from the state to the new state have  the Markov property:
stochastic transition probabilities between states depend only on the current state and action.
To model the RL problem as an MDP, at first, we represent a state which contains the features that are relevant to the decision of the agent. 
The transition between a state and an action to a new state is handled by the game engine. The aim of the agent is to select the best possible action in a state according to some performance metric, also called the reward function.

In RL there must be some function that maps a state (or state-action pair)
to a reward, also called the reward function. The aim of the agent in RL is to maximize the total expected reward that the agent receives in the long run through this reward function, also called the gain ($G$): 
\begin{equation}
    G = \sum_{t=0}^\infty r_t \cdot \gamma^t
\end{equation}
$r_t$ indicates the reward the agent receives at time $t$ and $\gamma$ is the discount factor. This discount factor is a number between 0 and 1 and controls how much future rewards are discounted and therefore how much immediate rewards are preferred. 

\subsection{Q-learning and Experience Replay}
Q-learning \cite{Q-Learning} is the most often used RL algorithm and learns to predict the quality/Q-value of an action in a specific state. By iterating through all possible actions in state, the algorithm picks the action with the highest Q-value as the action that the agent should take in that state. The Q-value indicates how much reward in the long term, or how much gain, the agent can expect to receive when choosing action $a_t$ in state $s_t$ at time step $t$. This prediction is updated over time by shifting it towards the reward that the agent got for taking that action and the predicted value of the best possible action in the next state.

To predict the Q-value for an action in a state a multi-layer perceptron (MLP) is used, which is trained through backpropagation. The target for backpropagation for a state-action tuple $s_t,a_t$ of a transition $(s_t,a_t,r_t,s_{t+1})$ is:
%The tabular algorithm for the Q-learning value iteration update for a transition consisting out of a state $s_t$, an action $a_t$, a reward $r_t$, and a new state $s_{t+1}$ is:\\
%\[
%    Q(s_t,a_t) = Q(s_t,a_t) \cdot (1 - \alpha) + \alpha \cdot (r_t + \gamma \cdot \max_aQ(s_{t+1},a))  
%\]

%In this formula $\alpha$ indicates the learning rate. This formula is adapted so that it can be used to train an MLP by calculating the target for backpropagation for a state-action tuple $(s_t,a_t)$:\\
\[
    Target(s_t,a_t) = r_t + \gamma \cdot \max_aQ(s_{t+1},a)
    %\label{eq:Q-Target}
\]

%\subsubsection{Exploration}
It is necessary to explore the action space throughout training to avoid being stuck in local optima as much as possible. 
For Q-learning the $\epsilon$-greedy exploration \cite{RL:AI} was chosen due to its simplicity. 
The $\epsilon$ value indicates how likely it is that a random action is chosen, instead of choosing greedily the action with the highest Q-value. 
For this research the $\epsilon$ value is annealed exponentially from 1 to a specific value close to 0 over the course of training. This allows the agent to progress steadily while exploring alternative actions over the course of training.

%\subsubsection{Target Networks}
To stabilize Q-learning combined with artificial neural networks, Mnih et al. \cite{Mnih2013} introduced target networks. As the training of Q-learning maximizes over all possible actions taken in the next state, the combination of this training method with function approximators can lead to the deadly triad \cite{RL:AI}. This deadly triad leads to a high probability of the Q-function to diverge from the true value function over the course of training through a positive feedback loop. A possible remedy to this problem is Double-Q-learning \cite{doubleQ}, which uses two Q-value networks. For the training of one network, the other network is used to calculate the Q-value of the best action in the next state to avoid the positive feedback loop of the deadly triad. Mnih et al. simplify this approach by introducing a target network in addition to the Q-value network. The parameters of the Q-value network are copied to the target network sporadically. This way there is no need to introduce a new separate network, as the prediction of the maximal Q-values is still done by a slightly different network, mitigating the deadly triad effect.

%\subsubsection{Prioritized Experience Replay}
Lin \cite{Lin1992} introduced a technique named experience replay, which was also used by Minh et al. for DQN \cite{mnih2015} to further stabilize and improve the performance of Q-learning. In using experience replay every transition tuple $(s_t, a_t, r_t, s_{t+1})$ is stored in a buffer instead of being trained on directly. When this buffer reaches its maximum capacity the oldest transitions in it get replaced. To train the value network in every training step $N$ random transitions from the buffer are extracted. For each of these transitions the target for $(s_t,a_t)$ is calculated and then the value network is trained on this mini-batch.

One assumption of using backpropagation to train an MLP is that the samples that are used to train in the mini-batches are independent and identically distributed. This assumption does not hold for online Q-learning, as each new transition is a result of the previous transition. Therefore the random sampling from a large buffer of transitions partially restores the validity of this assumption. 

To improve experience replay, Schaul et al. \cite{Schaul2015} developed prioritized experience replay (PER). PER does not sample uniformly from the replay buffer, to transition $i$ it assigns the sampling probability: 
\[
    P(i) = \frac{{TDE}_i^\alpha}{\sum_k{TDE}_k^\alpha}
\]
In this formula $\alpha$ determines how much prioritization is used, where $\alpha=1$ would mean full prioritization.
TDE stands for the temporal difference error of transition $i$ and indicates how bad the network is at predicting transition $i$:
\[ 
    TDE_i = r_t + \gamma \cdot \max_aQ(s_{t+1},a) - Q(s_t,a_t)
\]
This implies that transitions which the value network is bad at predicting will be replayed more often, which was shown to lead to faster learning and better final performance in \cite{Schaul2015}. 

PER leads to more high TDE transitions to be replayed which skews the sample distribution, therefore Schaul et al. also introduce an importance sampling weight which decreases the magnitude of the weight change in the MLP for transition $i$ anti-proportionally to its $TDE_i$. A weight $w_i$ is applied to the weight changes induced by each transition $i$ of magnitude:
%This is done to reduce the bias of training on average on more high TDE transitions. Therefore a weight $w_i$ is applied to the weight changes induced by each transition $i$ of magnitude:
\[
w_i = (\frac{1}{N} \cdot \frac{1}{TDE_i})^\beta
\]
$N$ is the size of the sampled batch and $\beta$ controls the amount of importance sampling. In practice the weights are used in the Q-Learning update by multiplying the prediction error for transition $i$, used in backpropagation, by $w_i$.

\subsection{Continuous Action Space Learning Algorithms}
In many applications it is not possible to discretize the action space to a sufficient degree, either because fine-grained precision is needed in the actions or because the action space is so vast that discretization would lead to an enormous amount of possible actions. If discretization is not feasible, there is the need for the agent to output continuous actions. To achieve this, an approach in RL is to have an actor map a state to an action. The output of the actor for this research will be two numbers between 0 and 1, indicating the relative mouse position on the screen. A common method to train this actor is through an actor-critic algorithm \cite{prokhorov1997adaptive}. This method also includes a critic network alongside the actor network, which is a value network as in Q-learning. The critic predicts the value of states or state-action combinations and can therefore steer the actor towards choosing actions that will maximize the gain.

For exploration we apply Gaussian noise to the predicted action by the actor before the action is used in the game environment. The standard deviation of the Gaussian noise starts at 1 and exponentially decays towards a value close to zero over the course of training.

We will compare the three actor-critic learning methods Continuous Actor-Critic Learning Automaton (CACLA) \cite{CACLA}, Deterministic Policy Gradient (DPG) \cite{DPG}, and the new Sampled Policy Gradient (SPG) with Q-learning. DPG is an off-policy actor-critic algorithm which calculates the policy gradient of the actor of a state by backpropagation through the critic. CACLA is an on-policy actor-critic algorithm, which makes the actor more likely to take an action, if the action positively surprised the critic. SPG is an off-policy variation of CACLA. To train the actor it samples actions in the action space of a state and makes the best sampled action more likely to be predicted by the actor in that state. This offers the advantage of being able to use offline exploration for training. %The comparison between these actor-critic learning algorithms and Q-learning will show whether the higher precision in the action-space of the actor-critic algorithms outweighs their higher complexity.

As DPG and SPG are both off-policy algorithms, they can directly make use of prioritized experience replay. Even though CACLA is an on-policy algorithm, it can still be used with prioritized experience replay, which will be elaborated upon later. The TDE of the critics is taken as the priority value for each transition in PER. 

The target networks used for these algorithms are not updated fully every fixed number of steps. Instead the weights of the target network are slowly trailing behind the critic and the actor network, by shifting the weights a bit more towards the weights of the non-target networks in every step, as done in \cite{DDPG}. The $\tau$ parameter determines how quickly the weights of the target networks $w^{target}$ shift towards the real weight values $w^{real}$:
\[
    w^{target}_{t+1} = \tau \cdot w^{real} + (1 - \tau) \cdot w^{target}_t
\]

\subsubsection{Continuous Actor-Critic Learning Automaton (CACLA)}
CACLA as an actor-critic algorithm has both an actor and a critic. The actor is an MLP that gets a state representation as an input and outputs an action. The critic for CACLA is an MLP as well and also has the state representation as an input. The output of the critic is one value which indicates the total expected reward of the input state. Therefore this critic does not predict a Q-value of a state-action combination, it predicts how good a specific state $s_t$ is. The formula to train this critic on a transition tuple $(s_t, a_t, r_t, s_{t+1})$ using backpropagation is:%the formula \ref{eq:Q-Target} of Q-learning is modified into:
\[
    Target^{critic}(s_t) = r_t + \gamma * V(s_{t+1})
\]
In this formula $V(s)$ refers to the prediction of the CACLA critic of state $s$ and is trained with temporal difference learning \cite{Sutton:88}. It can be seen that the action $a_t$ is not incorporated into the update of the critic, so for the critic the action is treated as a part of the environment. Therefore the critic does not learn the optimal value function. It learns to predict the value function for the current policy, so it is on-policy.% more or less stochastic event in the environment. The algorithm is on-policy and the critic predicts the gain of states for the actor, not for

To train the actor, a critic is required to calculate the temporal-difference error (TDE) of a transition. The magnitude of the TDE indicates how surprised the critic is by the result of the action $a_t$ in state $s_t$ and the sign of the TDE indicates whether the critic is positively or negatively surprised:
\[
    TDE(s_t,r_t,s_{t+1}) = Target^{critic}(s_t) - V(s_t)
\]

If the critic was positively surprised by the action $a_t$ of transition $i$ then this action is made more likely to be predicted by the actor through backpropagation with the target:
\[
    \text{If }TDE(s_t,r_t,s_{t+1}) > 0: Target^{actor}(s_t) = a_t
\]

The on-policy algorithm CACLA also makes use of experience replay. 
For every transition, CACLA evaluates whether the action taken in that transition led to a better result than the result the critic would have expected the current policy to achieve. 
So even for transitions that are not generated by the policy that CACLA currently approximates, the algorithm learns correctly whether the current policy should be moved in the direction of the observed policy.

CACLA also has an extension named CACLA+Var \cite{CACLA}. This extension allows CACLA to train multiple times on a transition if that transition was especially positively surprising. CACLA+var keeps track of the running variance of the TDE:
\[
    var_{t+1} = (1 - \beta) \cdot var_t + \beta * \text{TDE}(s_t,r_t,s_{t+1})^2
\]
The number of updates of the actor for action $a_t$ of transition $i$ is then:
\[
    \text{Number of Updates} = ceil(\frac{TDE(s_t,r_t,s_{t+1})}{\sqrt{var_t}})
\]

\subsubsection{Deterministic Policy Gradient (DPG)}
Instead of having a value network as a critic, the DPG algorithm has a Q-value network. This network is trained in the same way as the network in Q-learning. The actor in DPG is equivalent to the actor in CACLA: it is an MLP that maps a state to an action.

DPG gets its name from its way of training the actor: the gradient of the policy (the actor) is calculated deterministically for a state $s_t$. Instead of analyzing if the action after state $s_t$ was good or better than expected, DPG uses backpropagation to calculate how to change the weights of the actor network in such a way that the output action of the actor will lead to a higher Q-value of the critic for state $s_t$. With $\alpha$ as the learning rate the backpropagation target for the actor network would be:
\[
    Target^{actor}(s_t) = \alpha \cdot \nabla_aQ(s,a)|_{s=s_t,a=\pi(s_t)} + \pi(s_t)
\]

We implement this by feeding the output of the actor directly into the critic to create a merged network. In this merged network the weights of the critic are frozen (cannot be changed) so that the critic won't change its weights to output a value as big as possible. This allows us to set the combined network a Q-value target which is a bit higher than what it currently predicts. The only way for this network to get closer to this target is by modifying the weights of the actor network. 

Changing the weights of the actor directly through the critic is a very fast and direct way to assess the local policy gradient.  However,
rapidly adapting to the critic in a hill-climbing manner has the disadvantage of potentially getting stuck in a local optimum.

\section{Sampled Policy Gradient (SPG)}
\label{SPG}

The idea of sampled policy gradient is simple: instead of calculating the weight changes of the actor deterministically through backpropagation, SPG samples actions from the action space for a specific state. 
These actions are made more likely to occur, if they are better than the action that the actor would predict currently for that state. 
The actor and critic network architectures in SPG are exactly the same as in DPG. 
On policy CACLA is limited to training on the actions taken in the transitions in the experience replay buffer, whereas SPG applies offline exploration to find a good action. 
Therefore, SPG is an off-policy version of CACLA, because it can probe any action of the action space, and move towards them if they are good.

To calculate the backpropagation target for the actor, SPG compares the evaluation of the action of a transition $Q(s_t,a_t)$  to the evaluation of the action predicted by the actor $Q(s_t,\pi(s_t))$. $\pi(s_t)$ denotes the actor prediction for state $s_t$:
\begin{equation}
    \text{If }Q(s_t,a_t) > Q(s_t,\pi(s_t)): Target(s_t) = a_t
\end{equation}

Additionally to using the action of a transition, SPG allows the application of any search algorithm to find sampled actions with a better evaluation than $Q(s_t,\pi(s_t))$. 

% Split below in parts and shorten
%To extend SPG from its pure offline-CACLA form this research studies 
SPG with offline Gaussian exploration (SPG-OffGE) further extends SPG. 
SPG-OffGE creates a new sampled action $S$ times by applying Gaussian noise to the best action found so far. 
If the evaluation of the new sampled action $Q(s_t, \text{sampled action})$ is better than the evaluation of the best action found so far $Q(s_t, \text{best action so far})$, then this sampled action becomes the new best action. 
Initially, the best action is the action predicted by the actor $\pi(s_t)$ and the first sampled action is the action $a_t$ of the transition $i$. 
If the evaluation of the best action is better than the evaluation of the action that the actor would take currently, then that best action is made more likely to be predicted by the actor in state $s_t$. 
The pseudocode for SPG-OffGE is given in Algorithm \ref{pseudocode:SPG-OffGE}.

The standard deviation of the Gaussian noise for offline exploration decreases exponentially towards a value close to zero during training. 
The lower the standard deviation of this noise, the more probable it is that the algorithm finds a better action than the current action, although this action could be only marginally better. 

\begin{algorithm}
    \begin{algorithmic}
         \STATE input = batch of N tuples $(s_t, a_t, r_t, s_{t+1})$

         \FOR{$(s_t, a_t)$ of transition $i$ in input}
            \STATE best $\leftarrow \pi(s_t)$
            \IF{$Q(s_t,a_t) > Q(s_t,\pi(s_t))$}
                     \STATE best $\leftarrow a_t$
            \ENDIF
            \FOR{count in 1,...,S}
                \STATE sampled $\leftarrow applyGaussian(\text{best})$
                \IF{$Q(s_t,\text{sampled}) > Q(s_t,\text{best})$}
                     \STATE best $\leftarrow$ sampled
                \ENDIF
            \ENDFOR
            \IF{$Q(s_t,best) > Q(s_t,\pi(s_t))$}
                \STATE $\text{Target}_i = \text{best}$
            \ENDIF
         \ENDFOR
        \end{algorithmic}
 \caption{Pseudocode for the target calculation of the actor training of the Sampled Policy Gradient Algorithm with offline Gaussian exploration (SPG-OffGE)}
 \label{pseudocode:SPG-OffGE}
\end{algorithm}

The performance of SPG heavily relies on an accurate critic. 
The critic in DPG needs to be accurate in the region of action space around the action that the actor currently predicts, $\pi(s_t)$, to calculate a correct policy gradient. 
Due to the Gaussian exploration around $\pi(s_t)$, the action space region is 
generally well explored, which makes the critic accurate in this region. 
The sampling of SPG can be non-local from $\pi(s_t)$, especially if it is supposed to escape local optima that DPG gets trapped in. 
Therefore, SPG requires extensive exploration and a good critic to maximize its performance.

Two additions of SPG-OffGE are investigated in this paper. 
Storing the Best Sampled Action found (SBA) in a transition is an offline exploration method that changes  the action from the last training iteration on a transition.
Online Gaussian Exploration (OnGE) uses sampling to select an action. 
Instead of just adding Gaussian noise to the output of the actor, the OnGE method samples $N$ times around the output of the actor using Gaussian noise. 
With enough samples, the OnGE approach approximates the maximization operation over all actions used in Q-learning. 
The critic evaluates the samples and then Gaussian noise is added to the sample with the best evaluation, which is then used as the action. 
This extension makes the action selection more greedy towards the critic evaluation and it could be used in any off-policy actor-critic algorithm.

%%%%%%%%%%%
% the game

\section{The Agar.io Game and RL}\label{sec:game}
Agar.io is a multi-player online game in which the player controls one or more cells. 
The game has a top-down perspective on the map of which the size of the visible area of the player is based on the mass and count of their cells. 
The goal of the game is to grow as much as possible by having the player's cell absorb food pellets, viruses, or other smaller enemy player's cells. 
The game itself has no end. Players can join an ongoing game at any point in time. 
Players start the game as a single small cell in an environment with other players' cells of all sizes. 
When all the cells of a player are eaten, that player loses and it can choose to re-enter the game.

Every cell in the game loses a small percentage of its mass in every time step. 
This makes it harder for large cells to grow quickly and it punishes inaction or hesitation. 
The game has simple controls. 
The cursor's position on the screen determines the direction all of the player's cells move towards. 
The player also has the option to 'split', in which case every player cell (given the cell has enough mass) splits into two cells of the same mass, both with half the mass of the original cells. 
One of these cells is shot in the direction of the cursor with a given momentum. 
Furthermore, the player has an option to have every cell 'eject' a small mass blob, which can be eaten by other cells or viruses.

The game, although having relatively simple core mechanics, has a complex and dynamic range of environments. 
The more massive a given cell is, the slower it moves. 
This forces players to employ strategies with long-term risks and rewards. 
A large player might decide to split in order to shoot a cell towards a small opponent which it might otherwise not be able to catch, but this makes them vulnerable to other large players who can eat the smaller cells. 
Similarly, a small player has to navigate an environment of larger players while avoiding getting cornered and eaten.

\begin{figure}[h]
    \centering
    \includegraphics[width=.45\textwidth]{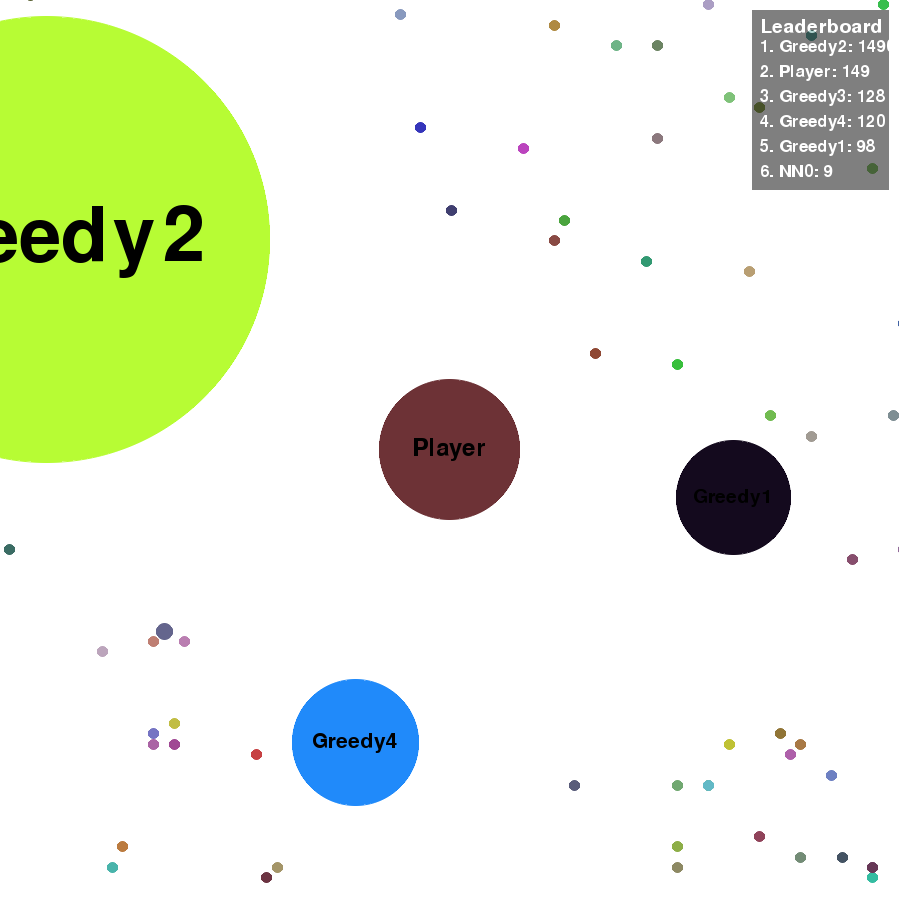}
    \caption{A clone of the game Agar.io used for this research. The player has one cell in the center of the screen. This player is in danger of being eaten by the Greedy bot seen on the top left, the other cells have a similar size as the player's cell and therefore pose no danger. The little colored dots are pellets that can be consumed to grow in mass.}
    \label{fig:agario}
\end{figure}

For the purpose of this research, the game was simplified to fit the available computational resources. 
The version of the game used has viruses disabled and is run with only one or two players. 
Furthermore, ejecting and splitting actions were disabled for the current experiments. 
Ejecting is only useful for very advanced strategies, and splitting requires tracking of when the player's cells are able to merge back together over long time intervals. 
The use of these actions would require recurrent neural networks such as LSTMs \cite{LSTM} which are outside of the scope of this research. 
Figure \ref{fig:agario} shows a screenshot of the clone of Agar.io used for this research.

Agar.io has the Markov property when the splitting functionality is disabled and the regular moving cells do not have any momentum and can change the direction of movement in any state, independent of previous states.
Solving the MDP is slightly modified for this research by using frame skipping \cite{ALE,Mnih2013}. 
In frame skipping a certain number of frames, or states, are skipped and a chosen action is repeated during all of these skipped frames, and the rewards during these skipped frames are summed up. 
Rewards are therefore larger and quickly propagated in state-action space. 
Frame skipping offers a direct computational advantage, as the agent does not have to calculate the best action in every frame and leads to successive states different from each other. 
Having greater differences between successive states makes it easier for a function approximator to differentiate them. 

%The aim in Agar.io is to grow as big as possible. 
An agent has the aim to maximize its combined overall mass of all its cells in the shortest amount of time possible in order to grow as big as possible. 
This leads to the idea of the reward being the change in mass between the previous state and the current state. 
If the agent dies, then the punishment is equivalent to the agent's mass one step before dying. 
Dying is not completely equivalent to just losing mass, as the agent also has to go through the struggle of starting completely anew. 
Thus, dying has a fixed punishment. 
But, dying with a large mass is also worse than dying with a low mass. 
While these two reward adjustments are not necessary for an agent to learn to play the game, the agent has a more human-like play style, as it fears death more, especially when it is very big. 
The reward function for the agent with $m_t$ being the mass at time $t$ is the following:
\[ r_t = 
    \begin{cases}
        m_t - m_{t-1},& \text{if alive}_t\\
        (m_{t-1} \cdot (-1.4)) - 40, & \text{otherwise}
    \end{cases}
 \]

An alternative to giving the change in mass as a reward is to simply give the agent the total cell mass in a state as a reward. 
Preliminary tests showed that this reward function also enables the agent to learn, but we opted for the change in mass as a reward function because this reward function is more likely to induce natural exploration. 

We compare the RL agents against each other and introduce a 'Greedy' bot to the game. 
This bot is preprogrammed to move towards the cell with the highest $\frac{\text{cell mass}}{\text{distance to greedy bot}}$ ratio. It ignores cells with a mass above its biggest own cell's absorption threshold. The bot also has no splitting or ejecting behavior. This relatively naive heuristic, outperforms human players at early stages of the game. On the other hand, the heuristic can be outperformed by abusing its lack of path planning and general world knowledge later in the game.

\subsection{The Action and State Representation}\label{sec:state}
The continuous action RL algorithms output two values between 0 and 1 that determine the relative mouse position where the player's cells will move to. However, as Q-learning iterates over all possible actions in a state, the action space cannot be continuous. 
Therefore we discretize the action space by laying a grid over the screen. 
Every center point of a square in the grid indicates a possible mouse position that the algorithm can choose. 
For example if the grid size per side is 2 the possible actions are: (0.25, 0.25), (0.75, 0.25), (0.25, 0.75), (0.75, 0.75). 
The resolution used in this paper is 5x5, resulting in 25 actions. 
Preliminary results showed that resolutions of 3x3 to 10x10 perform similarly well.

It has been shown by DeepMind \cite{mnih2015}, Lample \cite{lample2017}, and others that learning from pure pixel input is possible even in non-trivial environments. 
The convolutional neural network (CNN) learns to extract the relevant state features itself.
However, by using a CNN, the artificial neural network grows in size, number of hyperparameters, and general complexity. 
This leads to much longer training times and requires higher processing power per training step. 
On top of that, if we use pixel input for an image of size 84x84 with three color channels, we get 84x84x3 floating point numbers of around 4 bytes. 
For DQN \cite{mnih2015} and DDPG \cite{DDPG} experience replay buffer sizes of $10^6$ are used, of which every single experience contains two states. 
So we get $84 \cdot 84 \cdot 3 \cdot 4 \cdot 2 \cdot 10^6 \approx 1.69 \cdot 10^{11}$ bytes $ \approx 169$ Gigabytes. 
Unfortunately, nowadays most people do not have access to 170 Gigabytes of RAM required by this computation.

Instead, we make use of semantic vision grids similarly to Knegt et al. \cite{knegt2017}. 
These vision grids contain semantic information that could be extracted by hand-crafted preprocessing. 
For pure pellet collection only one grid for the pellets is used. 
When the agent trains against enemies, two more grids are added: a grid for enemy cells and a grid for walls. 
All grids have a size of 11x11 and cut the field of view of the agent into 121 equally sized areas. 
Each area of a grid is represented by one floating point number, therefore with an experience replay buffer of size $10^6$ and with enemies, one needs at least $11 \cdot 11 \cdot 3 \cdot 4 \cdot 2 \cdot 10^6 \approx 2.70$ Gigabytes. 
This would allow for even finer grids than 11x11.

Next to these vision grids two scalar values are used in the state representation: the total mass of the agent's cells and the size of the field of view of the agent relative to the size of a single pellet. The former value is necessary for the RL algorithm to predict the reward, but it could be replaced by a vision grid of the agent's cells. The latter value is not essential, but it did improve performance in preliminary experiments.

\begin{figure}
    \centering
    \includegraphics[width=.5\textwidth]{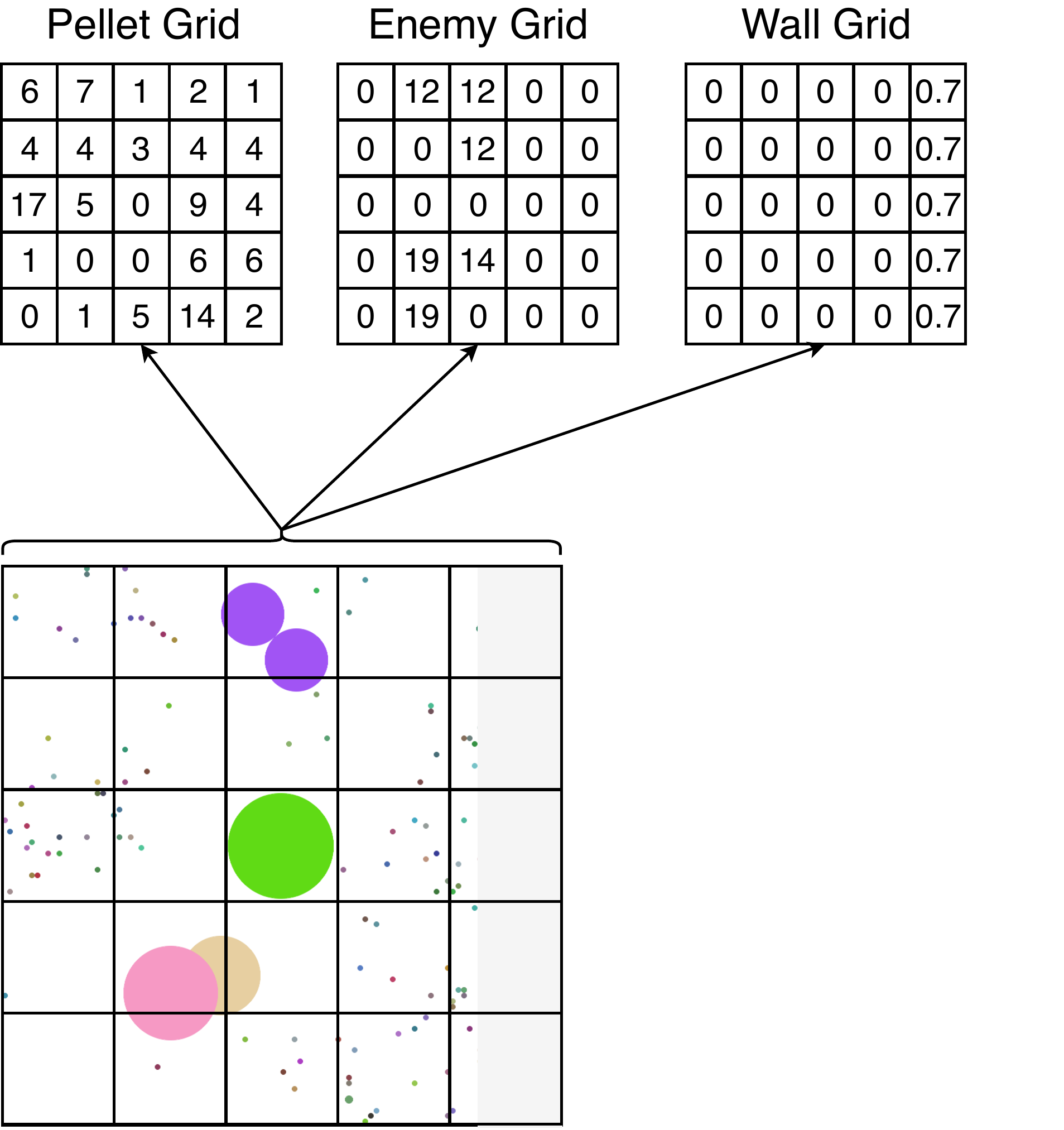}
    \caption{This example shows grids of size 5x5, the actual used grid size is 11x11. The semantic vision grid state representation consists of a grid laid out on the player's view. Values are extracted for different semantic objects. For instance, the pellet grid on the left counts the total pellet mass for every area of the grid.}
    \label{fig:grids}
\end{figure}

An illustration of a semantic vision grid representation can be seen in Figure \ref{fig:grids}. To calculate the floating point number of an area in the grid for pellets the total mass of the pellets in this area is summed. For the enemy grid the mass of the biggest enemy cell within that area is taken. The value for an area in the wall grid represents the fraction of area in that area which is covered by a wall.

The state representation in DQN also consists of images of the four last frames, but we only use the current frame. 
This is done because cells do not have momentum, and can change the direction of movement in every frame. 
Hence, information from previous states is mostly irrelevant. 
But, using previous frames might make it easier for the agent to deduce the moving speed and direction of other cells, and also there is most likely a correlation between the movement direction of an agent in concurrent frames. 
Therefore, a more extensive state representation might be advantageous, at the cost of requiring more computational power.

%A short discussion on how this semantic information can be acquired without access to the simulation engine can be found in Appendix \ref{ap:state}.

\section{Experiments and Results}
\label{sec:experiments}

\subsection{General Experimental Setup}
In this section, we describe three experiments we ran to compare the methods. 
The first experiment compares the different extensions of SPG. 
The second experiment tests the performance of all algorithms trained only with pellets, whereas the last experiment tests training the agents using self-play.

\subsubsection{Training}
There are two training environments:
\begin{enumerate}
    \item \textbf{Pellet collection:} In this type of experiment one agent is placed into an environment full of pellets, so the agent is only trained on collecting pellets. Only the pellet grid is used for the state representation.
    \item \textbf{Self-Play:} In this type of experiment two agents with the same learning algorithm are placed into an environment containing pellets. In every training step the learning algorithm is only applied once, whereas all the agents store their experienced transitions in the same replay buffer. The pellet grid, the wall grid, and the enemy grid are used for the state representation.
\end{enumerate}
In all experiments the environment is reset after 20,000 game steps. Upon reset all agents are reassigned a new cell with mass 10 at a random location and all pellet locations are randomized. This is done to avoid that the learning agents learn peculiarities of pellet locations on the map, and to force the agents to also learn to deal with low cell masses.

All algorithms were trained for 500,000 training steps, which equals 5,500,000 game steps, as the used frame skip rate was 10. On an Intel Xeon E5 2680v3 CPU @2.5Ghz it took Q-learning in the pellet collection environment approximately 7.5 hours to train, CACLA needed 12.5 hours, DPG needed 16.6 hours, SPG without sampling needed 17.2 hours, and SPG with 3 samples needed 29.3 hours (approximately 2.5-3 hours increase per sample).

\subsubsection{Testing}
\textbf{Test Environments:} While testing the noise factor of the agent ($\epsilon$ of $\epsilon$-greedy or standard deviation of Gaussian noise) is set to zero. The agent is placed into 1 or 2 environments to test its performance:
\begin{enumerate}
    \item \textbf{Pure Pellets:} In this environment the agent can only collect pellets for 15,000 steps. Every agent is tested in this environment.
    \item \textbf{Greedy Fight:} In this environment there is also a Greedy bot along with the agent and the pellets. The two bots fight for 30,000 steps, which is longer than the pellet environment time to enable bots with momentarily lower masses to catch up with the other bot. 
    Only agents that were trained with at least one other bot are tested in this environment.
\end{enumerate}

\textbf{Test Scores:} Every 5\% of the training the performance of one agent is tested five times per test environment. 
After training is completed the agents are placed in these two testing environments 10 times to measure the final post-training performance.

Furthermore, for testing the average performance of 10 runs for each environment is calculated by taking the mean of the performance scores of each run. The performance of an individual run is calculated by taking the mean mass that the agent had over the course of the whole run and by the highest mass that the agent achieved in a run. For each post-training run for one neural network initialization the maximum mass is taken, then over all the runs for that initialization the mean maximum mass is taken. Then again the mean of all mean masses of all initializations is calculated, along with the standard error.

\subsubsection{Algorithms}
The Q-learning architecture we use feeds only the state as an input to the network and has one output node per possible action, as used in \cite{Mnih2013}. An alternative would be to feed the action as an input as well and to have only one output node, but preliminary experiments showed that this is much slower and leads to worse performance.

We coarsely optimized all the hyperparameters using preliminary experiments. Initially for DPG and CACLA the parameters from the original papers \cite{DPG,CACLA} were used and later optimized.

All artificial neural networks/multi-layer perceptrons were constructed using Keras 2.1.4 \cite{chollet2015keras} and used Adam \cite{adam} as an optimizer. 
This research uses the OpenAi baselines repository \cite{baselines} for both uniform and prioritized experience replay. All used parameter values can be found in the Appendix.

\subsection{SPG Extensions}

\begin{figure}
    \centering
    \includegraphics[width=.8\textwidth]{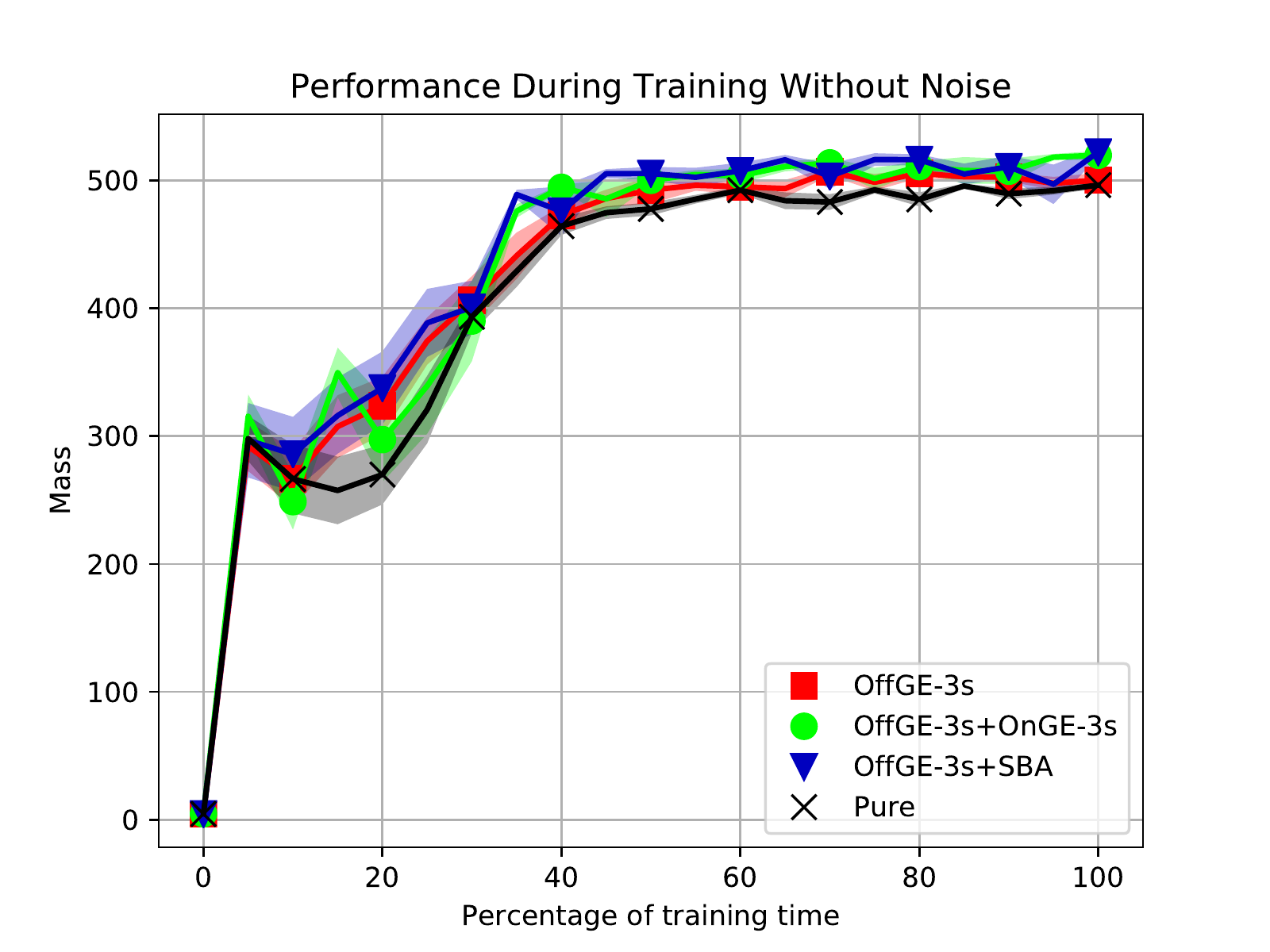}
    \caption{The performance (mass) over time of pure SPG and some extensions. The first extension is SPG with offline Gaussian exploration of 3 samples (OffGE-3s). Two other extensions are applied on top of this one: storing the best action (SBA) and online Gaussian exploration of 3 samples (OnGE-3s). The shaded regions mark the standard error deviations for every individual measurement. The performance of the algorithm was tested five times every 5\% of training. For each test the mean of the mean masses of 5 runs was taken.}
    \label{fig:pelletOverTimeSPG}
\end{figure}

To test which extensions of SPG are viable, all extensions plus the pure SPG were run on a pellet collection experiment. For the OffGE and OnGe extension three samples are used every training step, hence the name OffGE-3s and OnGE-3s. The results can be seen in Figure \ref{fig:pelletOverTimeSPG} and Table \ref{tab:pelletTrainingPelletTestSPG}.

\begin{table*}[!h]
\begin{center}
\scriptsize{
    \begin{tabular}{|c"c|c|c|c|}
    \hline
                            & \textbf{Mean Performance} & \textbf{StdError Mean} & \textbf{Max Performance} & \textbf{StdError Max} \\\thickhline
        \textbf{Pure}     & 456                      & 0.48                   & 642                     & 0.55                  \\\hline
        \textbf{OffGE-3s}     & 464                       & 0.74                   & 649                      & 0.89                  \\\hline
        \textbf{OffGE-3s+OnGE-3s} & 474              & 0.80                   & 661             & 0.70               \\\hline
        \textbf{OffGE-3s+SBA}      & \textbf{479}                       & 0.77                   & \textbf{667}                      & 0.93                  \\\hline
    \end{tabular}
    }
    \caption{Post-training performance of SPG and its extensions in the pellet collection task after being trained on only pellet collection.}
    \label{tab:pelletTrainingPelletTestSPG}
\end{center}
\end{table*}

It can be clearly seen that the pure SPG performs the worst and learns the slowest. The offline Gaussian exploration definitely improves performance. 
A t-test on the mean performance values between the offline Gaussian Exploration of 3 samples (OffGE-3s) and OffGE-3s with online Gaussian exploration of 3 samples (OnGE-3s) shows that there is a significant difference between the two with a p-value of 0.047. Tuning of the standard deviation and the number of samples used for online exploration might improve the performance of this extension even further.

A t-test on the mean performance values between OffGE-3s and OffGE-3s with storing the best action (SBA) shows that also here there is a significant difference between the two with a p-value of 0.005. This extension seems to definitely improve on the standard offline Gaussian exploration and is therefore recommended to be used as a default, especially because it does not introduce an additional tunable parameter.

\subsection{Pellet Collection}

\begin{figure}[h]
    \centering
    \includegraphics[width=.8\textwidth]{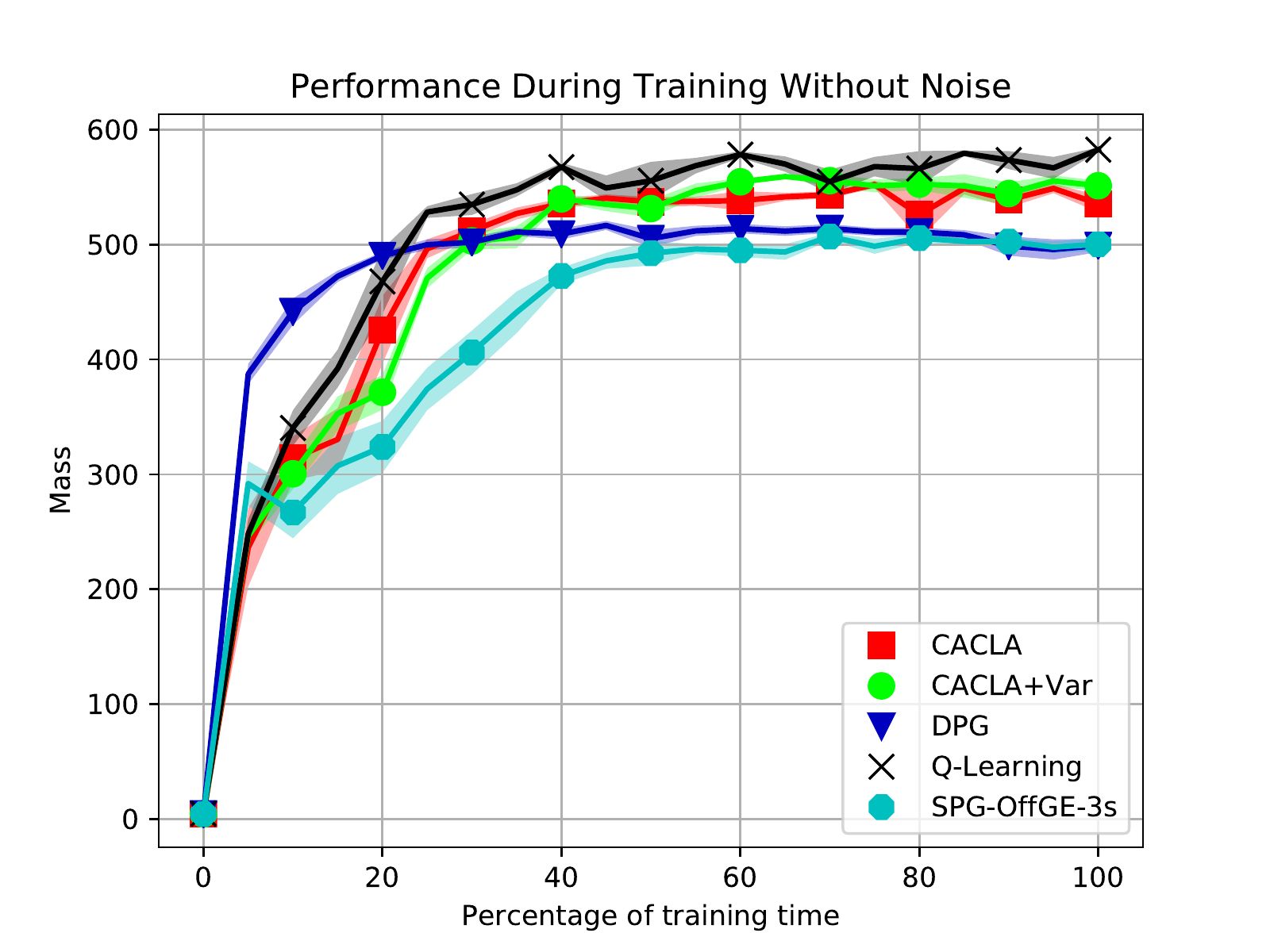}
    \caption{The pellet collection performance over time of all algorithms trained in a pure pellet environment}
    \label{fig:pelletOverTime}
\end{figure}

In this experiment the algorithms were only trained on pellet collection and are only tested on pellet collection. The results of the experiment are visible in Table \ref{tab:pelletTrainingPelletTest} and Figure \ref{fig:pelletOverTime}. The table shows the final mean and maximum performance along with the standard errors. The figure shows the performance of the algorithms during training.

\begin{table*}[!h]
\begin{center}
\scriptsize{
    \begin{tabular}{|c"c|c|c|c|}
    \hline
                            & \textbf{Mean Performance} & \textbf{StdError Mean} & \textbf{Max Performance} & \textbf{StdError Max} \\\thickhline
        \textbf{Random}     & 18.0                      & 0.12                   & 31.3                     & 0.17                  \\\hline
        \textbf{Greedy}     & 527                       & 0.33                   & 693                      & 0.57                  \\\hline
        \textbf{Q-Learning} & \textbf{528}              & 0.76                   & \textbf{753}             & 0.84                  \\\hline
        \textbf{CACLA}      & 496                       & 2.40                   & 699                      & 1.85                  \\\hline
        \textbf{CACLA+Var}  & 515                       & 0.55                   & 724                      & 1.29                  \\\hline
        \textbf{DPG}        & 460                       & 0.82                   & 649                      & 1.31                  \\\hline
        \textbf{SGP-OffGE-3s}    & 464                       & 0.74                   & 649                      & 0.89                  \\\hline
    \end{tabular}
    }
    \caption{Post-training performance in the pellet collection task after being trained on only pellet collection.}
    \label{tab:pelletTrainingPelletTest}
\end{center}
\end{table*}

It can be clearly seen that all algorithms outperform the Random bot, which takes a completely random action without skipping frames. Furthermore it can be seen that the algorithms Q-Learning, CACLA, and CACLA+Var outperform the Greedy bot in terms of maximal performance. The Greedy bot is very good at collecting pellets at the start as it does not use a frame skip rate and it can directly move towards the closest pellets. In the end game this strategy seems to not be optimal anymore, as the three learning algorithms outperform the Greedy method significantly.

When observing Figure \ref{fig:pelletOverTime} and Table \ref{tab:pelletTrainingPelletTest} the difference in performance of the algorithms becomes apparent. The algorithm with the highest final performance is Q-Learning, followed by CACLA+Var, and CACLA. DPG and SPG seem to have converged to an approximately similar final performance.

It is not surprising that Q-Learning does so well in this simple task: moving towards big clusters of pellets does not require precision. Therefore the extra complexity of the actor-critic algorithms that they need to get the precision in actions seems to hurt their performance in this task.

In Table \ref{tab:pelletTrainingPelletTest} it can be clearly seen that CACLA+Var significantly outperforms CACLA in terms of mean and maximal performance, further supporting the validity of this extension.

CACLA and CACLA+Var outperform the other actor-critic algorithms in this task. This might be the result of CACLA only needing to estimate how good a state is for its current policy, not how good an action is in a state in general. It can also make use of a simpler value network that does not take the action as an input, so it has a reduced complexity and seems to be able to make use of this to improve its policy even more.

It can be seen in Figure \ref{fig:pelletOverTime} that DPG converged very quickly, after 20\% of the training time the performance does not improve much more. As DPG calculates the exact policy gradient for the actor it is not surprising that it learns quicker than the other actor-critic algorithms. It can be assumed that SPG with perfect sampling of a small standard deviation would also find the steepest gradient to adjust the action, just as DPG. But other than DPG, SPG additionally has the ability to increase its sampling noise to escape the local optima that DPG quickly lands in.

\subsection{Self-Play}

\begin{figure}[h]
    \centering
    \includegraphics[width=.8\textwidth]{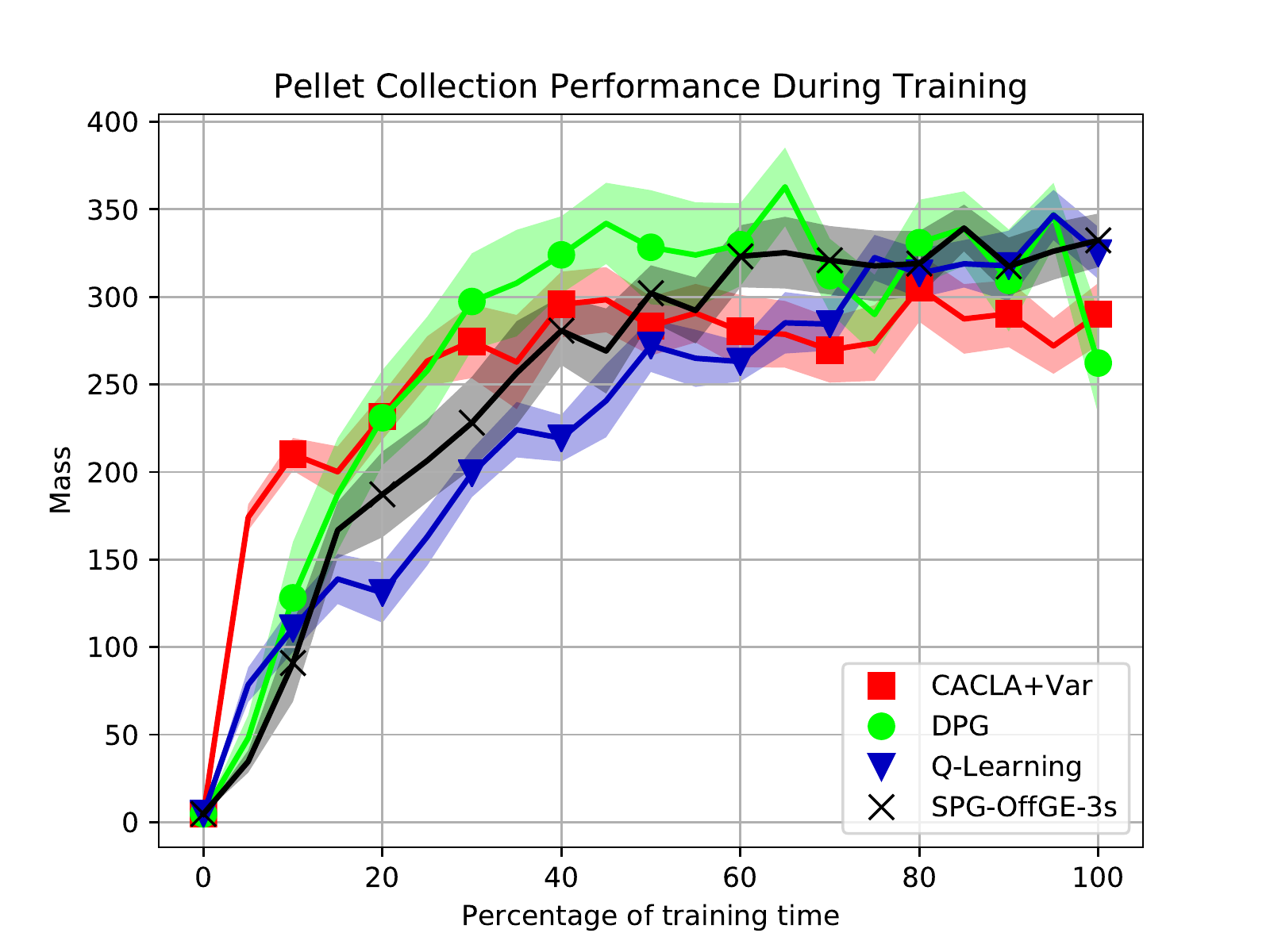}
    \caption{The pellet collection performance over time of the RL algorithms, trained in a 2 bot self-play scenario. }
    \label{fig:2NNPellet}
\end{figure}

\begin{figure}
    \centering
    \includegraphics[width=.8\textwidth]{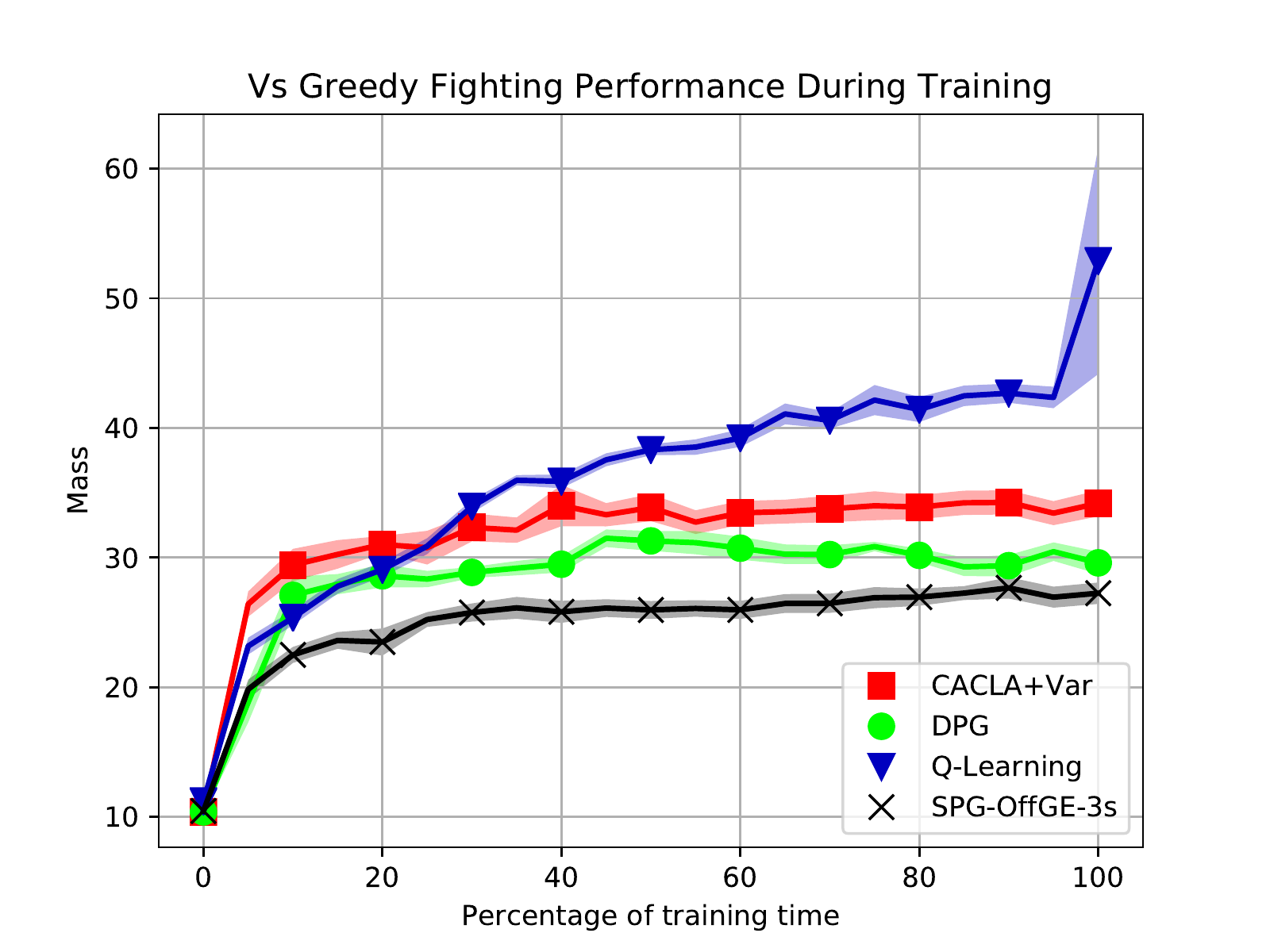}
    \caption{The fighting performance  versus the greedy bot over time of the RL algorithms, trained in a 2 bot self-play scenario.}
    \label{fig:2NNGreedy}
\end{figure}

In this experiment the algorithms CACLA+Var, Q-Learning, DPG, and SPG-OffGE-3s were trained in an environment consisting of two bots. Figure \ref{fig:2NNPellet} shows the performance score during training that the algorithms got in the pellet collection task and Figure \ref{fig:2NNGreedy} shows the performance score during training in the fighting task versus the greedy bot. Tables \ref{tab:2NNPellet} and \ref{tab:2NNGreedy} show the corresponding post-training performances for both tasks.

\begin{table*}[!h]
\begin{center}
\scriptsize{
    \begin{tabular}{|c"c|c|c|c|}
    \hline
                            & \textbf{Mean Performance} & \textbf{StdError Mean} & \textbf{Max Performance} & \textbf{StdError Max} \\\thickhline
        \textbf{Random}     & 18.0                      & 0.12                   & 31.3                     & 0.17                  \\\hline
        \textbf{Greedy}     & 527                       & 0.33                   & 693                      & 0.57                  \\\hline
        \textbf{Q-Learning} & 327              & 2.92                   & \textbf{507}             &    3.49               \\\hline
        \textbf{CACLA+Var}  & 284                       & 6.21                   & 449                      & 7.30                  \\\hline
        \textbf{DPG}        & 279                       & 10.5                   & 446                      & 15.5                  \\\hline
        \textbf{SGP-OffGE-3s}    & \textbf{335}                       & 5.95                   & 460                      & 6.57                  \\\hline
    \end{tabular}
    }
    \caption{Post-training performance in the pellet collection task, after being trained in self-play of two bots.}
    \label{tab:2NNPellet}
\end{center}
\end{table*}

\begin{table*}[!h]
\begin{center}
\scriptsize{
    \begin{tabular}{|c"c|c|c|c|}
    \hline
                            & \textbf{Mean Performance} & \textbf{StdError Mean} & \textbf{Max Performance} & \textbf{StdError Max} \\\thickhline
        \textbf{Random}     & 19.2                      & 0.13                   & 41.3                     & 0.34                  \\\hline
        \textbf{Greedy}     & 551                       & 12.8                   & 820                      & 13.5                  \\\hline
        \textbf{Q-Learning}      & \textbf{44.5}                       & 0.29                   & \textbf{142}      & 1.15                  \\\hline
        \textbf{CACLA+Var}  & 34.4                       & 0.32                   & 104                      & 1.48                  \\\hline
        \textbf{DPG}        & 30.0                       & 0.22                   & 97.6                      & 1.10                  \\\hline
        \textbf{SGP-OffGE-3s}    & 27.1                       & 0.25                   & 95.5                      & 0.66                  \\\hline
    \end{tabular}
    }
    \caption{Post-training performance in the fighting task versus the greedy bot, after being trained in self-play of two bots.}
    \label{tab:2NNGreedy}
\end{center}
\end{table*}

Again, all the algorithms are clearly better than random, as can be seen in the tables. But no algorithm comes close to the performance of the greedy bot in either pellet collection or fighting. The algorithms still learn to collect pellets, but generally much worse than when trained only with pellets. 

The performance against the greedy bot is very low: in testing the bots get nearly constantly eaten by the greedy bot. This is not very surprising: the greedy strategy is very good at the beginning of the game, so if the greedy bot has the upper hand in terms of mass early on, then it can repeatedly eat the other small bot. 

When observing the behavior of the bots it can be seen that they do learn to flee from the greedy bot, but during the chase they get trapped in the corner. To properly flee from a larger bot in a corner it is necessary to move strictly in parallel of a wall in one direction. This requires the bot to repeat one action for a large number of steps, but while doing so the greedy bot comes closer, triggering an earlier acquired "instinct" of moving in the opposite direction of the greedy bot. Therefore to solve this problem we propose that a better exploration mechanism is required which either acts upon some intrinsic motivation or which acts on larger time frames than individual actions.

When comparing the algorithms it can be seen that the same order in final performance occurs as in the pure pellet training: Q-learning is best, followed by CACLA+Var, and DPG and SPG attain a similar performance. This time CACLA+Var learns very quickly, probably due to its state-value network it can directly associate a negative reward with the presence of another bot. SPG seems to be learning to collect pellets better than DPG or CACLA+Var, but fails to outperform them on the more important task of fighting the greedy bot.

Overall, the discretization of the action space leads to very good results with Q-learning, outperforming all other algorithms in all tasks. 
More complicated tasks involving fighting requires more sophisticated algorithms with better exploration.

%\subsection{General Discussion}
In reinforcement learning research, many implementations of the same algorithm give different results \cite{RLthatMattersHenderson}. 
Minor implementational details, which are not reported in the research papers might have a large influence on performance.
The learning algorithms used in this paper are constructed using the Keras library \cite{chollet2015keras} and the experience replay buffer from OpenAi \cite{baselines}. 
This was done to make sure that no algorithm uses special tricks over another that would give an algorithm the edge.

All algorithms have been tuned as well as possible in a limited time and limited computation through a coarse search through hyperparameter space.
But, the optimality of the hyperparameters for an algorithm is not guaranteed. 
The current hyperparameters of the algorithm seem to be in a local optimum, as small perturbations to single parameters do not lead to improvements, but there might of course be better hyperparameter values.

\section{Conclusions}\label{sec:conclusions}

Our study on how to apply reinforcement learning to the game Agar.io has led to a new off-policy actor-critic algorithm named Sampled Policy Gradient (SPG). 
We compared some state of the art actor-critic continuous action RL algorithms,  CACLA and DPG, to SPG and Q-learning on the Agar.io game.

The results show that Q-learning and CACLA perform very well in the pellet collection task by outperforming a preprogrammed Greedy bot that follows a simple, yet effective, heuristic.
The other two algorithms perform slightly worse than the Greedy bot in pellet collection, but still significantly better than random. 
All algorithms were not able to successfully learn how to defeat the Greedy bot heuristic in a direct fighting scenario, requiring further research.
The analysis indicates that CACLA performed well in comparison to the other actor-critic algorithms because of its use of a state-value critic function. 
The DPG algorithm learns very quickly but gets stuck in a local optimum. SPG learns slower than DPG, but catches up with DPG during training to reach the same final performance.

The novel SPG algorithm was shown to result in similar final performances as DPG, even without heavy extensions. 
An online Gaussian sampling method improves the performance of SPG. 
Storing the best sampled action in the transition to use it in later training instances increased performance significantly. 
We recommend to use offline Gaussian exploration, with the method of storing the best sampled action, as a default version of SPG.

For the relatively simple task of pellet collection, Q-learning was able to
obtain the best performances, despite that it cannot directly handle
continuous actions and needed to discretize the action space.
It was shown that the lack in precision through discretization of the action space is balanced out by a more simple and better learning network.
 
\subsection{Future Work}
First of all it is important to note that only a large simplification of Agar.io was investigated. Future work could use more sophisticated algorithms such as critics with distributional value output, hierarchical actor-critic architectures, and exploration by curiosity. To deal with the time dependency that the cell splitting functionality brings, LSTM units \cite{LSTM} can be used instead of perceptrons.

The analysis of the SPG algorithm shows how the DPG algorithm can be matched in performance by some simple sampling in action space. Future work could explore the algorithm space between DPG and SPG further by using the policy gradient from DPG to steer the sampling of SPG.
%
%As CACLA+Var was shown to outperform pure CACLA, a natural extension to SPG could be SPG+Var. Instead of the variance of the temporal difference error this algorithm would keep track of the variance of the difference between $Q(s_t,\pi(s_t))$ and $Q(s_t,\text{best sampled action})$. 

It was shown that CACLA performs better than the other actor-critic algorithms. It was hypothesized that this is because CACLA uses a state value function instead of a state-action value function. Therefore it would be interesting to compare QV-Learning \cite{QV} applied on SPG and DPG to CACLA. If QV-Learning improves the performance of the critic, then it might drastically improve the performance of SPG and DPG, which are both heavily reliant on an accurate critic.
%
%To escape local optima the offline exploration noise of SPG can also be dynamic. One possible extension of the algorithm could keep track of how many transitions of a batch it succeeded in finding a better action. If this fraction is large then the exploration noise is too low and should be increased and vice versa. Such dynamic noise adaptations could boost SPG's performance drastically, as its ability to escape local optima is properly made use of.
%

Lastly it is of great importance that SPG is tested in other environments against CACLA and DPG. The surprising success of CACLA over DPG and SPG might not replicate in environments in which it is more difficult to estimate the value of a state or in environments with a sparse reward function. Also DPG might perform better than SPG in other environments, even though the work of this paper suggests that the SPG algorithm matches the performance of DPG and allows for a wide 
range of extensions.

\section*{Acknowledgements}

We would like to thank the Center for Information Technology of the University of Groningen for their support
and for providing access to the Peregrine high performance computing cluster.

\bibliographystyle{plain}
\bibliography{literature}

\begin{thebibliography}{10}

\bibitem{ALE}
M.~G. {Bellemare}, Y.~{Naddaf}, J.~{Veness}, and M.~{Bowling}.
\newblock {The Arcade Learning Environment: An Evaluation Platform for General
  Agents}.
\newblock {\em ArXiv e-prints}, July 2012.
\newblock arxiv:1207.4708.

\bibitem{chollet2015keras}
Fran\c{c}ois Chollet et~al.
\newblock Keras.
\newblock \url{https://keras.io}, 2015.

\bibitem{Cybenko1989}
G.~Cybenko.
\newblock Approximation by superpositions of a sigmoidal function.
\newblock {\em Mathematics of Control, Signals and Systems}, 2(4):303--314, Dec
  1989.

\bibitem{baselines}
P.~Dhariwal, C.~Hesse, O.~Klimov, A.~Nichol, M.~Plappert, A.~Radford,
  J.~Schulman, S.~Sidor, and Y.~Wu.
\newblock Open{AI} baselines.
\newblock \url{https://github.com/openai/baselines}, 2017.

\bibitem{doubleQ}
Hado~V. Hasselt.
\newblock Double {Q}-learning.
\newblock In J.~D. Lafferty, C.~K.~I. Williams, J.~Shawe-Taylor, R.~S. Zemel,
  and A.~Culotta, editors, {\em Advances in Neural Information Processing
  Systems 23}, pages 2613--2621. Curran Associates, Inc., 2010.

\bibitem{RLthatMattersHenderson}
P.~{Henderson}, R.~{Islam}, P.~{Bachman}, J.~{Pineau}, D.~{Precup}, and
  D.~{Meger}.
\newblock {Deep Reinforcement Learning that Matters}.
\newblock {\em ArXiv e-prints}, September 2017.
\newblock arxiv:1709.06560.

\bibitem{LSTM}
Sepp Hochreiter and J\"{u}rgen Schmidhuber.
\newblock Long {S}hort-{T}erm {M}emory.
\newblock {\em Neural Comput.}, 9(8):1735--1780, November 1997.

\bibitem{adam}
D.~P. {Kingma} and J.~{Ba}.
\newblock {Adam: A Method for Stochastic Optimization}.
\newblock {\em ArXiv e-prints}, December 2014.
\newblock arxiv:1412.6980.

\bibitem{knegt2017}
Stefan J.~L. Knegt, Madalina~M. Drugan, and Marco Wiering.
\newblock Opponent modelling in the game of tron using reinforcement learning.
\newblock In {\em International Conference on Agents and Artificial
  Intelligence}, 2018.

\bibitem{lample2017}
Guillaume Lample and Devendra~Singh Chaplot.
\newblock Playing {FPS} games with deep reinforcement learning.
\newblock In {\em AAAI}, pages 2140--2146, 2017.

\bibitem{DDPG}
T.~P. {Lillicrap}, J.~J. {Hunt}, A.~{Pritzel}, N.~{Heess}, T.~{Erez},
  Y.~{Tassa}, D.~{Silver}, and D.~{Wierstra}.
\newblock {Continuous control with deep reinforcement learning}.
\newblock {\em ArXiv e-prints}, September 2015.
\newblock arxiv:1509.02971.

\bibitem{Lin1992}
Long-Ji Lin.
\newblock Self-improving reactive agents based on reinforcement learning,
  planning and teaching.
\newblock {\em Machine Learning}, 8(3):293--321, May 1992.

\bibitem{Mnih2013}
V.~{Mnih}, K.~{Kavukcuoglu}, D.~{Silver}, A.~{Graves}, I.~{Antonoglou},
  D.~{Wierstra}, and M.~{Riedmiller}.
\newblock {Playing {A}tari with Deep Reinforcement Learning}.
\newblock {\em ArXiv e-prints}, December 2013.
\newblock arxiv:1312.5602.

\bibitem{mnih2015}
Volodymyr Mnih, Koray Kavukcuoglu, David Silver, Andrei~A Rusu, Joel Veness,
  Marc~G Bellemare, Alex Graves, Martin Riedmiller, Andreas~K Fidjeland, Georg
  Ostrovski, et~al.
\newblock Human-level control through deep reinforcement learning.
\newblock {\em Nature}, 518(7540):529, 2015.

\bibitem{prokhorov1997adaptive}
Danil~V Prokhorov and Donald~C Wunsch.
\newblock Adaptive critic designs.
\newblock {\em IEEE transactions on Neural Networks}, 8(5):997--1007, 1997.

\bibitem{Schaul2015}
T.~{Schaul}, J.~{Quan}, I.~{Antonoglou}, and D.~{Silver}.
\newblock {Prioritized Experience Replay}.
\newblock {\em ArXiv e-prints}, November 2015.
\newblock arxiv:1511.05952.

\bibitem{shantia2011}
A.~{Shantia}, E.~{Begue}, and M.~A. {Wiering}.
\newblock Connectionist reinforcement learning for intelligent unit micro
  management in {S}tarcraft.
\newblock In {\em Neural Networks (IJCNN), The 2011 International Joint
  Conference on}, pages 1794--1801. IEEE, 2011.

\bibitem{DPG}
David Silver, Guy Lever, Nicolas Heess, Thomas Degris, Daan Wierstra, and
  Martin Riedmiller.
\newblock Deterministic policy gradient algorithms.
\newblock In {\em Proceedings of the 31st International Conference on
  International Conference on Machine Learning - Volume 32}, ICML'14, pages
  I--387--I--395. JMLR.org, 2014.

\bibitem{Sutton:88}
Richard~S. Sutton.
\newblock Learning to predict by the methods of temporal differences.
\newblock {\em Machine Learning}, 3(1):9--44, 1988.

\bibitem{RL:AI}
Richard~S. {Sutton} and Andrew~G. {Barto}.
\newblock {\em Reinforcement Learning: An Introduction}.
\newblock MIT Press, 2017.

\bibitem{tesauro1995}
Gerald Tesauro.
\newblock Temporal difference learning and {TD}-{G}ammon.
\newblock {\em Communications of the ACM}, 38(3), 1995.

\bibitem{CACLA}
H.~{Van Hasselt} and M.~A. {Wiering}.
\newblock {Reinforcement Learning in Continuous Action Spaces}.
\newblock {\em Proceedings of the 2007 IEEE Symposium on Approximate Dynamic
  Programming and Reinforcement Learning (ADPRL 2007)}, 2007.

\bibitem{Q-Learning}
C.~J. C.~H. {Watkins}.
\newblock {\em Learning from Delayed Rewards}.
\newblock PhD thesis, King's College, Cambridge, 1989.

\bibitem{QV}
M.~Wiering.
\newblock {QV ($\lambda$)-learning: A new on-policy Reinforcement Learning
  Algorithm}.
\newblock In D.~Leone, editor, {\em Proceedings of the 7th European Workshop on
  Reinforcement Learning}, pages 17--18, 2005.

\end{thebibliography}

\newpage

\section*{Appendix: Parameter values}

\scalebox{1}{
    \begin{tabular}{l|l}
        \textbf{Parameter}                                    & \textbf{Value}        \\ \hline
        Reset Environment After                               & 20,000 training steps \\
        Frame Skip Rate                                       & 10                    \\
        Discount Factor                                       & 0.85                  \\
        Total Training Steps                                  & 500,000               \\
        Optimizer                                             & Adam                  \\
        Loss Function For All Algorithms                      & Mean-Squared Error    \\
        Semantic Vision Grid Size                             & 11x11                 \\
        Weight Initializer                                    & Glorot Uniform        \\
        Activation Function Hidden Layers All Algorithms      & ReLU                 \\
        Prioritized Experience Replay Alpha                   & 0.6                   \\
        Prioritized Experience Replay Beta                    & 0.4                   \\
        Prioritized Experience Replay Capacity                & 75,000                \\
        Training Batch Length                                 & 32                    \\
        Q-Learning Steps Between Target Network Updates       & 1500
        \\
        Q-Learning Hidden Layers                              & 3                     \\
        Q-Learning Neurons per Hidden Layer                   & 256                   \\
        Q-Learning Learning Rate                              & 0.0001                \\
        Q-Learning Std of Noise at Training End               & 0.0004                \\
        Q-Learning Number of actions                          & 5x5                   \\
        Actor-Critic Policy Output Activation Function        & Sigmoid               \\
        Actor-Critic Actor and Critic Hidden Layers           & 3                     \\
        Actor-Critic Critic Neurons per Hidden Layer          & 250                   \\
        Actor-Critic Actor Neurons per Hidden Layer           & 100                   \\
        CACLA Target Network $\tau$                              & 0.02                  \\
        CACLA Actor Learning Rate                             & 0.0005                \\
        CACLA Critic Learning Rate                            & 0.000075              \\
        CACLA+Var Start Variation                             & 1                     \\
        CACLA+Var Beta                                        & 0.001                 \\
        DPG Target Network $\tau$                                & 0.001                 \\
        DPG Actor Learning Rate                               & 0.00001               \\
        DPG Critic Learning Rate                              & 0.0005                \\
        DPG Combined Network Q-Value Target Increase ($\beta$) & 2                     \\
        DPG Feed Action in Layer                              & 1                     \\
        DPG Critic L2 Weight Decay                            & 0.001                 \\
        SPG Target Network $\tau$                                & 0.001                 \\
        SPG Actor Learning Rate                               & 0.0005                \\
        SPG Critic Learning Rate                              & 0.0005                \\
        SPG Offline Sample Number                             & 3                     \\
        SPG Offline Exploration Noise at End of Training      & 0.0004                \\
        SPG Online Sample Number                              & 3                     \\
    \end{tabular}
    }

%\clearpage

%\begin{appendices}
%    \input{tex/appendixA.tex}
%\end{appendices}

\end{document}